\documentclass[10pt,twocolumn,letterpaper]{article}

\usepackage{cvpr}      % To produce the REVIEW version
\usepackage{wrapfig}
\usepackage{xspace}
\usepackage{pifont}
\usepackage{utfsym}
\usepackage{enumitem}

\newcommand{\latinabbrev}[1]{#1.\xspace}

\usepackage{multirow}
\usepackage{colortbl}

\definecolor{cvprblue}{rgb}{0.21,0.49,0.74}
\usepackage[pagebackref,breaklinks,colorlinks,allcolors=cvprblue]{hyperref}

\usepackage{fancyhdr}
\def\paperID{} % *** Enter the Paper ID here
\def\confName{CVPR}
\def\confYear{2025}

\title{{\fontsize{12.7pt}{14pt}\selectfont XQ-GAN: An Open-source Image Tokenization Framework for Autoregressive Generation}}

\def\eg{\latinabbrev{e.g}}

\def\ie{\latinabbrev{i.e}}

\definecolor{verylightgray}{RGB}{234, 250, 254}
\definecolor{fgreen}{RGB}{15, 159, 94}

\author{%
  Xiang Li$^{1*}$, Kai Qiu$^{1*}$, Hao Chen$^1$, Jason Kuen$^2$, Jiuxiang Gu$^2$, Jindong Wang$^3$, Zhe Lin$^2$, Bhiksha Raj$^{1}$ \\
  $^1$Carnegie Mellon University, $^2$Adobe Research, $^3$William \& Mary\\
}

\begin{document}

\twocolumn[{
\renewcommand\twocolumn[1][]{#1}
\twocolumn[
\maketitle]
\vspace{-1cm}
\begin{center}
    \renewcommand{\arraystretch}{1.2} % 调整表格行间距
    \setlength{\tabcolsep}{8pt} % 调整列间距
    \scalebox{0.85}{
    \begin{tabular}{p{4.5cm}
<{\centering}p{2.5cm}
<{\centering}|p{2.cm}
<{\centering}p{2.5cm}
<{\centering}|p{2.5cm}
<{\centering}p{2.87cm}
<{\centering}}
\hline
Quantization & Reference & Backbone & Reference & Alignment & Reference \\
\hline
Vector Quant. (VQ) & VQVAE \cite{van2017neural} & CNN & VQGAN \cite{esser2021taming} & DINO-v2 & ImageFolder \cite{li2024imagefolder} \\
Residual Quant. (RQ) & RQVAE \cite{lee2022autoregressiveimagegenerationusing} & Transformer & ImageFolder \cite{li2024imagefolder} & CLIP & VILA-U \cite{wu2024vila} \\
\cline{3-6}
Product Quant. (PQ) & ImageFolder \cite{li2024imagefolder} & Discriminator & Reference & Pre-training & Reference \\
\cline{3-6}
Look-up Free Quant. (LFQ) & MAGVIT-v2 \cite{yu2024language} & StyleGAN & StyleGAN \cite{karras2019style} & ImageNet & ImageNet \cite{deng2009imagenet} \\
Binary Spherical Quant. (BSQ) & BSQ-ViT \cite{zhao2024image} & PatchGAN & PatchGAN \cite{isola2017image} & LAION-400M & LAION-400M \cite{schuhmann2021laion} \\
Multi-Scale RQ (MSRQ) & VAR \cite{tian2024visualautoregressivemodelingscalable} & DINO & VAR \cite{tian2024visualautoregressivemodelingscalable} & IMed-361M & IMed-361M \cite{cheng2024interactive} \\
\hline
\end{tabular}
}
\vspace{-0.3cm}
\captionof{table}{
    Overview of XQ-GAN components and their design choices. The framework consists of five main components: Encoder, Decoder, Quantization, Semantic alignment, and Discriminator, each offering multiple design choices for exploration. We provide pre-trained weights on several datasets and support fine-tuning with full parameters, LORA \cite{hu2021lora} and frozen encoder/decoder.
    % We list the leveraged design choices in the reference.
}
\label{tab:xqgan_components}
\end{center}
}
]

\begin{abstract}
Image tokenizers play a critical role in shaping the performance of subsequent generative models. Since the introduction of VQ-GAN, discrete image tokenization has undergone remarkable advancements. Improvements in architecture, quantization techniques, and training recipes have significantly enhanced both image reconstruction and the downstream generation quality. In this paper, we present \textbf{XQ-GAN}, an image tokenization framework designed for both image reconstruction and generation tasks. Our framework integrates state-of-the-art quantization techniques, including vector quantization (VQ), residual quantization (RQ), multi-scale residual quantization (MSVQ), product quantization (PQ), lookup-free quantization (LFQ), and binary spherical quantization (BSQ), within a highly flexible and customizable training environment. On the standard ImageNet 256×256 benchmark, our released model achieves an rFID of 0.64, significantly surpassing MAGVIT-v2 (0.9 rFID) and VAR (0.9 rFID). Furthermore, we demonstrate that using XQ-GAN as a tokenizer improves gFID metrics alongside rFID. For instance, with the same VAR architecture, XQ-GAN+VAR achieves a gFID of 2.6, outperforming VAR’s 3.3 gFID by a notable margin. To support further research, we provide pre-trained weights of different image tokenizers for the community to directly train the subsequent generative models on it or fine-tune for specialized tasks.
\url{https://github.com/lxa9867/ImageFolder}
\end{abstract}
\section{Introduction}
Autoregressive (AR) image generation \cite{tian2024visualautoregressivemodelingscalable,li2024controlvar,wu2024vila,weber2024maskbit} has achieved notable progress and demonstrated promising performance recently. 
The current advanced image generation paradigm typically requires a pre-trained image tokenizer \cite{esser2021taming,li2024imagefolder,yu2024language,yu2024spae,yu2024imageworth32tokens} that encodes the image into a more compact latent space, where an AR generator, \eg vision transformer \cite{alexey2020image}, is leveraged to model the latent distribution, for both efficiency and effectiveness in generation \cite{rombach2022high}.
% . 
% After that, 
% on which an autoregressive generator, \eg vision transformer \cite{alexey2020image}, is leveraged to model the latent distribution. 

\begin{figure}[t]
    \centering
    \includegraphics[width=\linewidth]{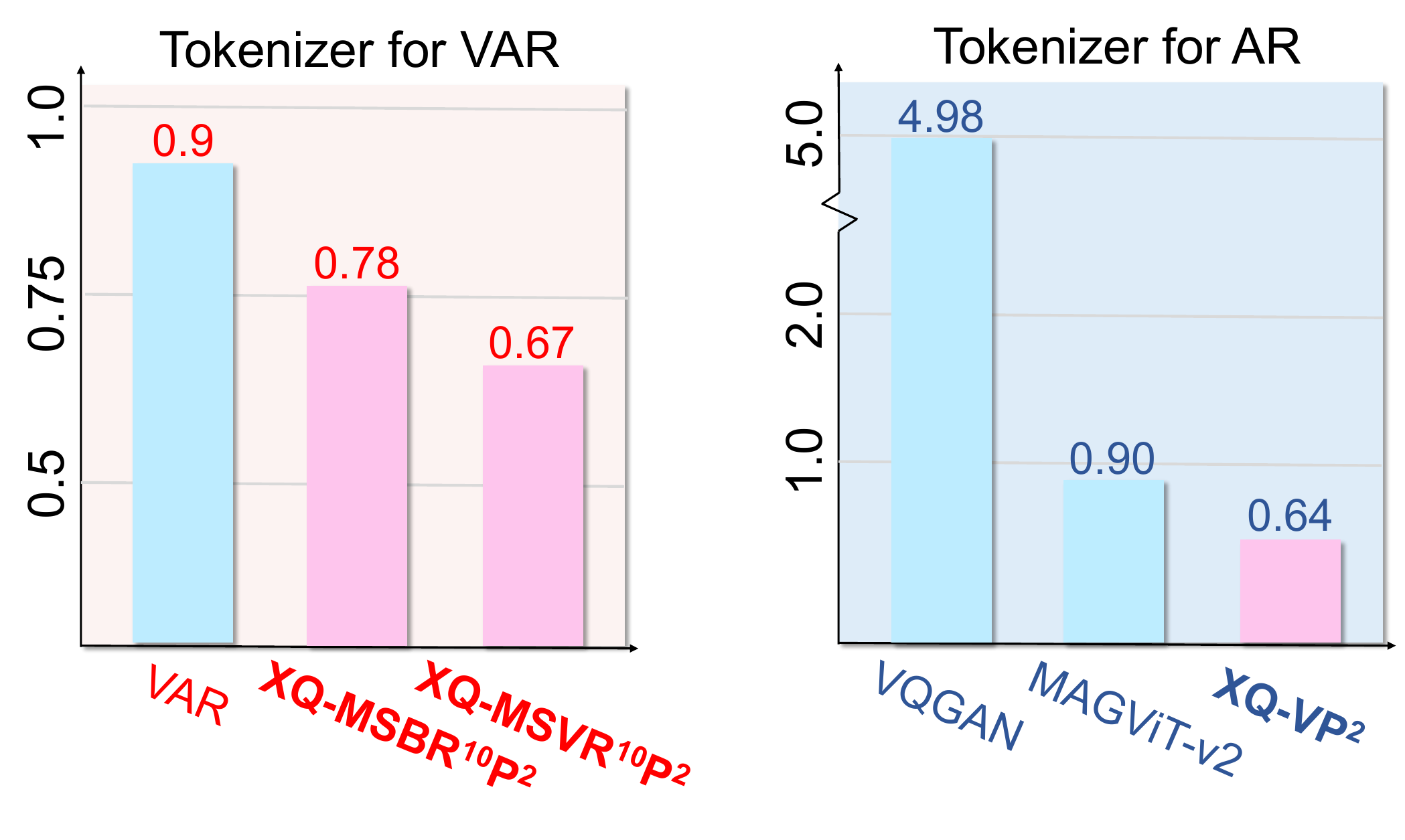}
    \vspace{-0.6cm}
    \caption{Performance comparison of XQGAN and prior arts on ImageNet 256x256 reconstruction benchmark. We provide XQGAN's tokenizer for vanilla AR and VAR \cite{tian2024visualautoregressivemodelingscalable} modeling. Our proposed XQGAN demonstrates superior performance against mainstream tokenizers for both AR and VAR tasks. XQ-GAN variants are named with XQ-\{multi-scale (MS)\}-\{vector quantization (V), lookup free quantization (L), binary spherical quantization (B)\}-\{residual quantization (R)\}$^{N}$-\{product quantization (P)\}$^{P}$ where $N$ and $P$ denotes residual depth and product branch number respectively.
    }
    \label{fig:teaser}
\end{figure}

Recent approaches \cite{li2024imagefolder,wu2024vila,sun2024autoregressive} demonstrate that the construction of latent space has a significant impact on the subsequent image generation quality. 
Different variables, such as token number \cite{li2024imagefolder,sun2024autoregressive}, semantics \cite{li2024imagefolder,wu2024vila,tian2024visualautoregressivemodelingscalable} and codebook utilization \cite{yu2021vector,zhu2024scaling,yu2024language} have been shown to significantly affect the generation quality in previous research.
% are variables that impact the generation quality significantly concluded from previous researches. 

For AR generation, the quantization approaches are vital to learn the latent representation. 
VQVAE \cite{van2017neural} introduces a vector quantization (VQ) approach by leveraging the idea of the nearest neighbor to encode continuous tokens into a finite set of discrete tokens. After that, residual quantization (RQ) \cite{lee2022autoregressiveimagegenerationusing} and multi-scale residual quantization (MSRQ) \cite{tian2024visualautoregressivemodelingscalable} are further proposed to hierarchically quantize the continuous tokens. More recently, product quantization \cite{li2024imagefolder} (PQ) is introduced to separate the continuous space into several smaller subspaces and separately quantize them. Look-up free quantization (LFQ) \cite{yu2024language} and its improved version binary spherical quantization (BSQ) \cite{zhao2024image} addressed the codebook utilization issue when handling large codebook size.

Recent studies \cite{li2024imagefolder,yu2024representation} also indicate that a semantic-rich representation can substantially benefit the generation quality. 
From the tokenizer side, ImageFolder \cite{li2024imagefolder} leverages DINO-v2 \cite{oquab2023dinov2} to inject semantics into the product quantized tokens. Similarly, VILA-U \cite{wu2024vila} utilizes CLIP \cite{radford2021learning} to enhance the semantics of image tokens for unified understanding and generation using a large language model (LLM). SoftVQ-VAE \cite{Chen2024softvqvae} explores several different semantic alignment combinations and manages to achieve a highly compact representation, \ie, 32 tokens, for both reconstruction and generation.
However, there is no unified framework for training tokenizers and subsequent generation tasks with different techniques proposed in previous methods.

In this paper, we propose XQ-GAN, an open-source and highly customizable framework for image tokenization and subsequent generation.
The proposed framework presents several advantages over the previous ones \cite{luo2024open,sun2024autoregressive}.
\begin{itemize}[leftmargin=1em,nosep]
\setlength\itemsep{0em}
    \item As shown in \cref{tab:xqgan_components}, XQ-GAN provides implementations of various quantization approaches, encoder and decoder backbones, discriminator architectures, and semantic alignment approaches, 
    thus permitting a modular and flexible combination of them.
    \item To further facilitate the community, we provide several pre-trained weights on large-scale datasets, such as ImageNet \cite{deng2009imagenet} (1000-class natural image), LAION-400M \cite{schuhmann2021laion} (text-rich natural image) and IMed-361M \cite{cheng2024interactive} (14 categories, multimodal medical image). 
    \item Following the standard benchmarking setting, we benchmark our framework on several settings. Notably, the best performance of the proposed XQ-GAN significantly outperforms the previous counterparts MAGVIT-v2 and VAR by a large margin.
\end{itemize}
\section{Related Works}
\subsection{Image Tokenizers}
Image tokenizer has seen significant progress in multiple image-related tasks. Traditionally, autoencoders \cite{hinton2006reducing,vincent2008extracting} have been used to compress images into latent spaces for downstream work such as (1) generation and (2) understanding. In the case of generation, VAEs \cite{van2017neural,razavi2019generating} learn to map images to probabilistic distributions; VQGAN \cite{esser2021taming,razavi2019generatingdiversehighfidelityimages} and its subsequent variants \cite{lee2022autoregressive,yu2023language,mentzer2023finite, zhu2024scaling,takida2023hq,huang2023towards,zheng2022movqmodulatingquantizedvectors,yu2023magvit,weber2024maskbit,yu2024spae,luo2024open,zhu2024addressing} introduce a discrete latent space for better compression for generation. On the other hand, understanding tasks, such as CLIP \cite{radford2021learning}, DINO \cite{oquab2023dinov2,darcet2023vitneedreg,zhu2024stabilize}, rely heavily on LLM \cite{vaswani2023attentionneed,dosovitskiy2021imageworth16x16words} to tokenize images into semantic representations \cite{dong2023peco,ning2301all}. These representations are effective for tasks like classification \cite{dosovitskiy2021imageworth16x16words}, object detection \cite{zhu2010deformable}, segmentation \cite{wang2021maxdeeplabendtoendpanopticsegmentation}, and multi-modal application \cite{yang2024depth}. For a long time, image tokenizers have been divided between methods tailored for generation and those optimized for understanding. After the appearance of \cite{yu2024an}, which proved the feasibility of using LLM as a tokenizer for generation, some works \cite{wu2024vila} are dedicated to unify the tokenizer for generation and understanding due to the finding in \cite{gu2023rethinkingobjectivesvectorquantizedtokenizers}. 
% However, because of the different perspectives in such two tasks: generation focuses on learning the high-frequency information from images to maintain fidelity, while understanding concentrates more on semantic information. To address these challenges and unify the tokenizer effectively, our work introduces a novel approach that employs \textbf{product quantization} to create a representation capable of two tasks. By balancing the high-frequency details needed for generation with the semantic information critical for understanding, our method enables a more efficient tokenization process. 

\subsection{Autoregressive Visual Generation}
Autoregressive models have shown remarkable success in generating high-quality images by modeling the distribution of pixels or latent codes in a sequential manner. Early autoregressive models such as PixelCNN \cite{van2016conditional} pioneered the approach of predicting pixel values conditioned on previously generated pixels. The transformer architecture \cite{vaswani2023attentionneed}, first proposed in NLP, has spread rapidly to image generation \cite{shi2022divaephotorealisticimagessynthesis, mizrahi20244m} because of its scalability and efficiency. MaskGIT \cite{chang2022maskgitmaskedgenerativeimage} accelerated the generation by predicting tokens in parallel, while MAGE \cite{li2023magemaskedgenerativeencoder} applied MLLM \cite{bao2022beitbertpretrainingimage, peng2022beitv2maskedimage} to unify the visual understanding and the generation task. Recently, autoregressive models continued to show their scalability power in larger datasets and multimodal tasks \cite{he2024mars}; models like LlamaGen \cite{sun2024autoregressive} adapting Llama \cite{touvron2023llamaopenefficientfoundation} architectures for image generation. New directions such as VAR \cite{tian2024visualautoregressivemodelingscalable,li2024controlvar}, MAR \cite{li2024autoregressiveimagegenerationvector} and Mamba \cite{li2024scalable} have further improved flexibility and efficiency in image synthesis. Currently, more and more unified multimodal models like SHOW-O \cite{xie2024show}, Transfusion \cite{zhou2024transfusion} and Lumina-mGPT \cite{liu2024lumina} continue to push the boundaries of autoregressive image generation, demonstrating scalability and efficiency in diverse visual tasks.

\begin{figure*}[t]
    \centering
    \includegraphics[width=\linewidth]{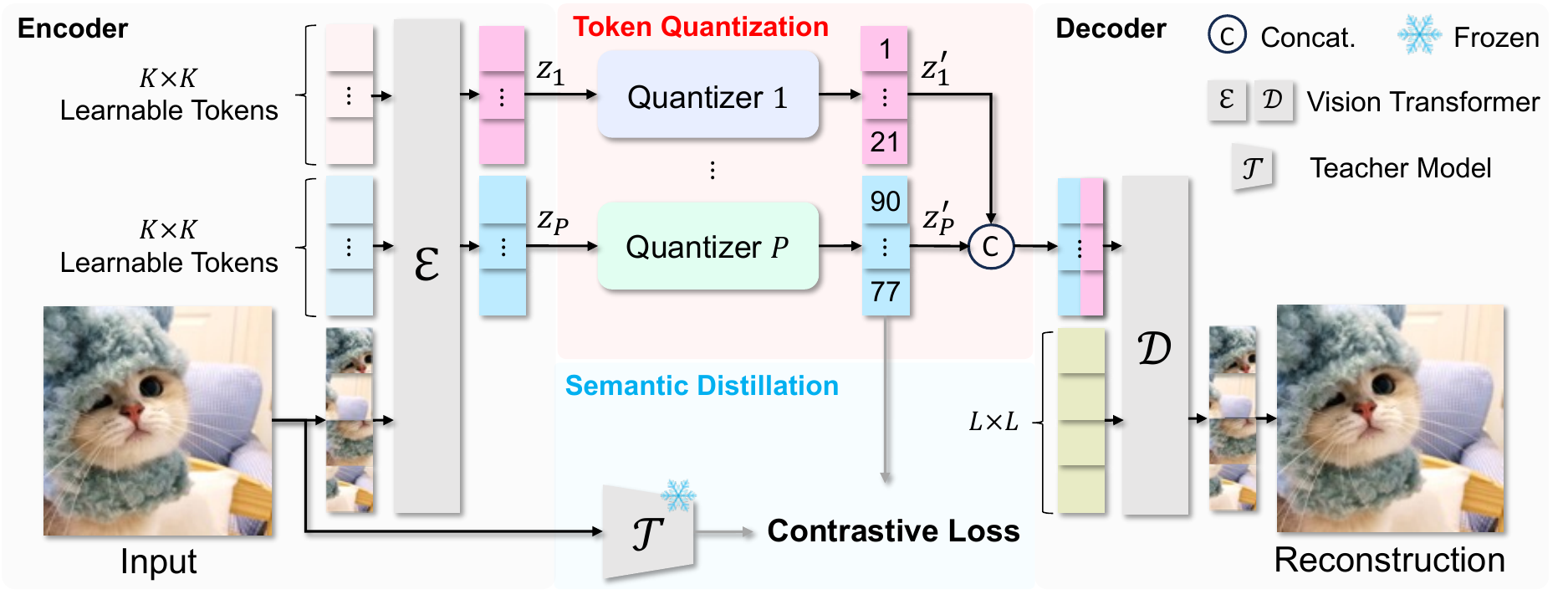}
    \vspace{-0.7cm}
    \caption{Overview of XQ-GAN-SC pipeline with latent spatial compression (SC). In this pipeline, a vision transformer is adopted as the encoder and decoder. $P$ set of $K\times K$ learnable tokens are utilized as queries to query the image tokens. $P$ denotes the quantizer number where $P=1$ is the vanilla Vector Quantization \cite{esser2021taming} setting and $P>1$ denotes a Product Quantization \cite{li2024imagefolder} setting. $K\times K$ denotes the spatial resolution of the quantized latent. During decoding, $L\times L$ learnable tokens are leveraged to query the quantized tokens and then decoded to the reconstructed image. The cost of image decoding is independent to the quantizer number, making it suitable for generation tasks that only require decoding during inference.}
    \vspace{-0.2cm}
    \label{fig:XQ-GAN-spatial}
\end{figure*}

\subsection{Diffusion Models for Image Generation}
Diffusion models, initially introduced by Sohl-Dickstein et al. \cite{sohldickstein2015deepunsupervisedlearningusing} as a generative process and later expanded into image generation by progressively infusing fixed Gaussian noise into an image as a forward process. A model, such as U-Net \cite{ronneberger2015u}, is then employed to learn the reverse process, gradually denoising the noisy image to recover the original data distribution. In recent years, this method has witnessed significant advancements driven by various research efforts. Nichol et al. \cite{nichol2021improveddenoisingdiffusionprobabilistic}, Dhariwal et al. \cite{dhariwal2021diffusionmodelsbeatgans}, and Song et al. \cite{song2022denoisingdiffusionimplicitmodels} proposed various techniques to enhance the effectiveness and efficiency of diffusion models, paving the way for improved image generation capabilities. Notably, the paradigm shift towards modeling the diffusion process in the latent space of a pre-trained image encoder as a strong prior \cite{van2017neural,esser2021taming} rather than in raw pixel spaces \cite{vahdat2021scorebasedgenerativemodelinglatent,rombach2022highresolutionimagesynthesislatent} has proven to be a more efficient and instrumental method for high-quality image generation. Moreover, a lot of research on the model architecture replaces or integrates the vanilla U-Net with a transformer \cite{peebles2023scalablediffusionmodelstransformers} to further improve the capacity and efficiency of multi-model synthesis on diffusion model. Inspired by these promising advancements in diffusion models, numerous foundational models have emerged, driving innovation in both image quality and flexibility. For instance, Glide \cite{nichol2021glide} introduced a diffusion model for text-guided image generation, combining diffusion techniques with text encoders to control the generated content. Cogview \cite{ding2021cogview} leveraged transformer architectures alongside diffusion methods to enhance image generation tasks. Make-A-Scene \cite{gafni2022make} and Imagen \cite{saharia2022photorealistic} focused on high-fidelity image synthesis conditioned on textual inputs, showcasing the versatility of diffusion models across modalities. DALL-E \cite{ramesh2021zero} and Stable Diffusion \cite{rombach2022high} brought diffusion models to mainstream applications, demonstrating their ability to generate high-resolution photorealistic images.  
\section{XQ-GAN}
XQ-GAN is a customizable framework for advanced image tokenization. Two types of pipelines are implemented - XQ-GAN-Spatial Compression (SC), for reducing the latent resolution, and XQ-GAN-Vanilla (V), for being compatible with classic tokenizers. Semantic alignment and adversarial discriminators can be further applied to these pipelines.

\begin{figure}[t]
    \centering
    \includegraphics[width=\linewidth]{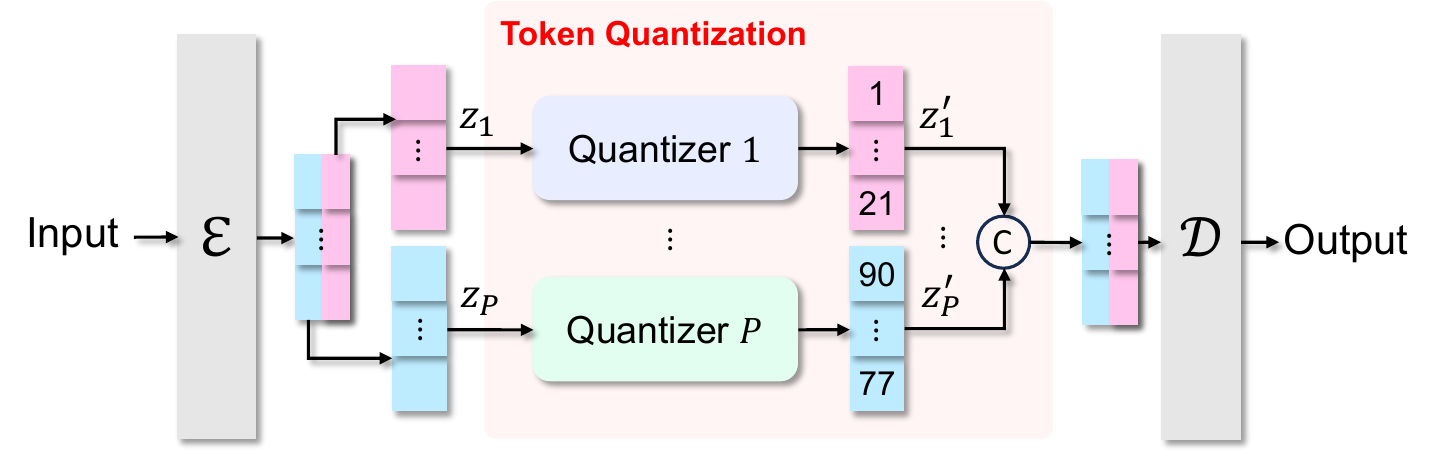}
    \vspace{-0.6cm}
    \caption{XQ-GAN-V pipeline.}
    \vspace{-0.3cm}
    \label{fig:XQ-GAN-V}
\end{figure}
\begin{figure*}[t]
    \centering
    \includegraphics[width=\linewidth]{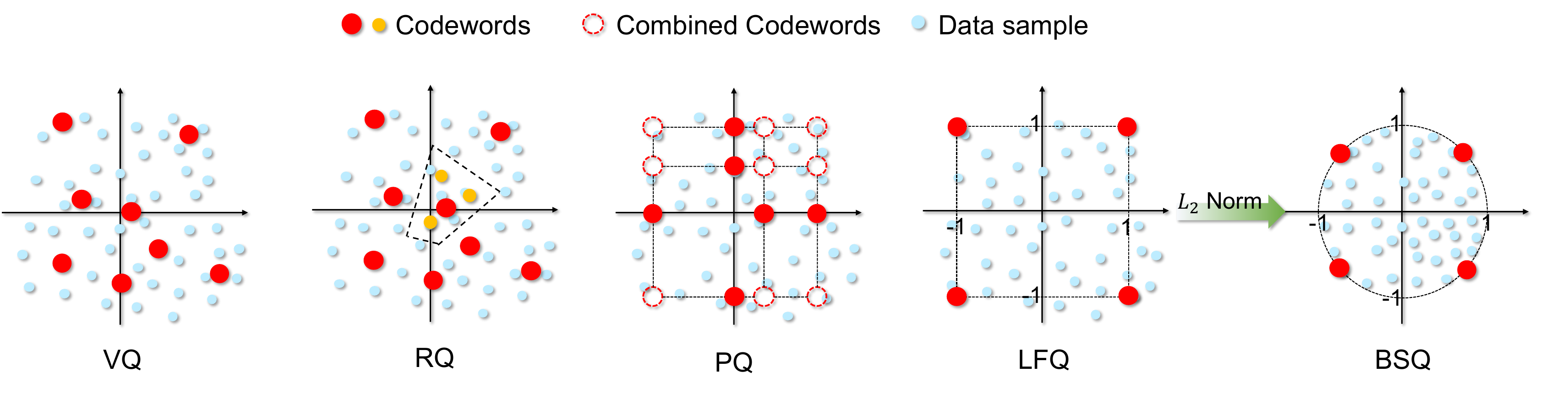}
    \vspace{-0.6cm}
    \caption{Visualization of Vector Quantization (VQ), Residual Quantization (RQ), Product Quantization (PQ), Lookup Free Quantization (LFQ), and Binary Spherical Quantization (BSQ) in a simple two-dimensional space. VQ equivalent to k-means clustering, partitions the space into Voronoi regions based on the nearest centroids. RQ refines this region iteratively by quantizing residual at each step. PQ quantized the space with the combination of several codewords on subspaces. LFQ projects the tokens into several binary subspaces and quantizes each subspace with $[-1, 1]$. BSQ applied an $L_2$ normalization on LFQ's subspace prior to quantization, resulting in a spherical quantization boundary.}
    \label{fig:quantization}
\end{figure*}
\subsection{XQ-GAN-SC}
The XQ-GAN-SC pipeline tokenizes images with three main components - Encoder, Decoder, and Token Quantization module. Following ImageFolder \cite{li2024imagefolder}, vision transformers are leveraged as the encoder and decoder. As shown in \cref{fig:XQ-GAN-spatial}, given an input image $I$, we first patchify it into a set of ${L\times L}$ tokens where $L$ is the patch size. After that, the image tokens are concatenated with $P$ sets of ${K\times K}$ learnable tokens and serve as the input to the transformer encoder $\mathcal{E}$. The same spatial positional encodings are added to the learnable tokens to inform the spatial alignment. A level embedding is additionally used to convey the difference across $K\times K$ tokens. Let us denote the encoded tokens corresponding to the learnable tokens as $z_1,\cdots,z_P$. To discretize them, we use different quantizers $\mathcal{Q}_1,\cdots,\mathcal{Q}_P$ to conduct product quantization ($P=1$ is equivalent to vanilla VQ). The quantized tokens $z^\prime_1,\cdots,z^\prime_P$ are further concatenated with a set of ${L\times L}$ learnable tokens to decode the reconstructed image with a decoder $\mathcal{D}$.

\subsection{XQ-GAN-V}

XQ-GAN-V pipeline follows the vanilla VQ-GAN \cite{esser2021taming} architecture while enhancing the token quantization part to support customizable quantization with different tokenization approaches. Specifically, as shown in \cref{fig:XQ-GAN-V}, given an input image $I$, we extract features with an encoder $\mathcal{E}$. After that, the extracted features are channel-wise chunked into several subtokens and separately conduct quantization. The quantized tokens $z^\prime_1,\cdots,z^\prime_P$ are further concatenated and fed to the decoder to reconstruct the image.

\section{Quantization}
\begin{figure}[t]
    \centering
    \includegraphics[width=\linewidth]{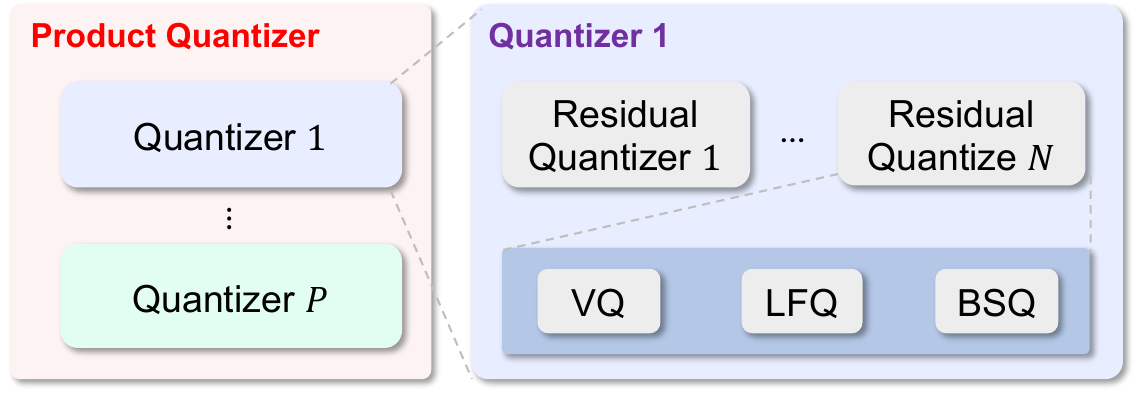}
    % \vspace{-0.6cm}
    \caption{Hierarchical quantizer design. With $P=1$, $N=1$, and VQ for residual quantizer, the quantizer is equivalent to the vanilla VQ \cite{esser2021taming}.}
    \label{fig:quantizer}
\end{figure}

To incorporate the combination of different quantization approaches, we propose a hierarchical pipeline. As shown in \cref{fig:quantizer}, XQ-GAN first decides the quantizer number $P$ for product quantization. After that, each quantizer is assumed to be a residual quantizer where each residual quantizer can be a vector quantizer, look-up free quantizer, or binary spherical quantizer. It is worth noting that, with a $P=1$, $N=1$, and VQ as residual quantizer, the framework is equivalent to conducting vanilla VQ proposed in VQGAN \cite{esser2021taming}.

\subsection{Vector Quantization}
Vector quantization (VQ) \cite{esser2021taming} introduces a learnable codebook that maps continuous input features to discrete latent representations, enabling generative models in a K-means manner. Specifically, each input continuous feature $x$ was processed with a learned, discrete codebook $\mathcal{C} =\{ e_j \}^J_{j=1}$ by mapping to its closest codeword $z$ as follow: 
\begin{equation}
    z^\prime=\arg\min_{e_j\in\mathcal{C}}\|z-e_j\|_2^2
\end{equation}
where $z^\prime$ is the quantized representation of the input.

\subsection{Residual Quantization}
Residual quantization (RQ) \cite{tian2024visualautoregressivemodelingscalable} leverages multiple tokens $\hat{z}_1, \hat{z}_2, \cdots, \hat{z}_N$ to represent an image, where $N$ denotes the number of residual steps. In the $i$-th residual step, the RQ quantizes the input $r_i$ by mapping it to its closest vector in the codebook $\mathcal{C}=\{e_j\}_{j=1}^J$ as $z_i^\prime = \arg\min_{e_j\in\mathcal{C}}\|r_i-e_j\|_2^2$. After the nearest-neighbor look-up, the quantized representation $z_i^\prime$ is subtracted by the input to form a residual representation which serves as the input to the next residual layer:

\begin{equation}
\begin{aligned}
    &z_{i} = z_{i-1}-z^\prime_{i-1}\\
    &r_i = r_{i-1} - z_i \\
    &z_0 = 0 \\
    &r_0 = 0 \\
\end{aligned}
\end{equation}
where $r$ is the residual. 

\paragraph{Quantizer dropout.}
Specifically, to enhance the representation capability, we notice that a quantizer dropout strategy \cite{kumar2024high,zeghidour2021soundstream} may improve generation performance.
During training, quantizer dropout randomly drops out the last several quantizers. The quantizer dropout happens with a ratio of $p$. With the quantizer dropout, the final output of the Residual Quantization can be formulated as:
\begin{equation}
    z^\prime = \sum_{i=1}^{n} z^\prime_i,  N_{start} \leq n \leq N.
\end{equation} 
Applying quantizer dropout enables residual quantizers to encode images into different bitrates depending on the residual steps.

\subsection{Product Quantization}
Product quantization (PQ) \cite{li2024imagefolder} aims to quantize a high-dimension vector to a combination of several low-dimension tokens, which has shown a promising capability to capture different contexts across sub-quantizers \cite{baevski2020wav2vec}. Given a continuous feature $z$, product quantization performs as:
\begin{equation}
    \mathcal{P}(z)=\mathcal{Q}_1(z_1)\oplus\cdots\oplus\mathcal{Q}_{P}(z_P), \mathcal{Q}_p\sim \mathcal{C}_p,
\end{equation}
where $\oplus$ denotes channel-wise concatenation (cartesian product), $\sim$ denotes that vector quantizer $\mathcal{Q}_p$ is associated with the codebook $\mathcal{C}_p=\{e_j\}_{j=1}^J$, i.e., $\mathcal{Q}_p$ maps a feature $z_p$ to a codeword $e_p=\arg\min_{e_j\in\mathcal{C}_p}\|z_p-e_j\|_2^2$ that minimizes the distance between $z_p$ and $e_j\in\mathcal{C}_p$. $P$ is the product number, \ie, sub-vector number and $z_p$ is a sub-vector from $z$ having $z=\oplus_{p=1}^Pz_p$. Product quantization first divides the target vector into several sub-vectors and then quantizes them separately. After the quantization, the quantized vectors resemble the original vector by concatenation.

\subsection{Lookup Free Quantization}
Lookup-free quantization (LFQ) \cite{yu2023magvit} is a special variant of product quantization that quantizes high-dimension vectors to a combination of several low-dimension tokens. LFQ decomposes the latent space as the Cartesian product of $\log_2^J$ single-dimensional variables and formulated as $\mathcal{C}=\times ^{\log_2^J}_{p=1} \mathcal{C}_p$ where $\mathcal{C}_p=\{1, -1\}$ and $J$ denotes the global codebook size. 
Given a feature vector $z\in \mathbb{R}^{\log_2^K}$, each dimension of the vector $z[p]$ can be separately quantized by a codebook $\mathcal{C}_p$, leading to a binary quantized value for each dimension, \ie, -1 or 1.

\subsection{Binary Spherical Quantization}
Similar to LFQ, binary spherical quantization (BSQ) has recently been proposed to improve its performance. Recall that the feature vector $z$ is defined in $\mathbb{R}^{\log_2^K}$, which may lead to substantial quantization error as $\mathcal{C}_p=\{-1,1\}$. This may barrier the straight-through gradient backpropagation \cite{Chen2024softvqvae}. In this way, BSQ additionally introduces a $L_2$ norm upon the feature vector $z$ to bound the quantization error to $[0,1]$.

\subsection{Multi-scale Residual Quantization}
VAR \cite{tian2024visualautoregressivemodelingscalable} enhances the vanilla RQ with a multi-scale setting. The feature map $r_{i-1}$ is first downsampled from its original dimension $L\times L$ to a lower resolution $L_i \times L_i$ where $L_i$ represents the resolution of current residual $r_i$. After downsampling, the look-up procedure is performed to match each feature vector with the closest codebook vector. After the look-up, the quantized vector $z_i$ is upsampled back to the original dimension $L\times L$ to ensure consistency across scales. Due to the loss of high-frequency information from downsampling, \cite{tian2024visualautoregressivemodelingscalable} utilized a 2D convolutional layer after upsampling to restore the lost details and enhance the fidelity of the reconstructed image. Specifically, this convolutional layer processes the upsampled feature vectors according to the equation:
\begin{equation}
    \hat{z}_i^\prime = \gamma\times\text{conv}(z_i^\prime) + (1-\gamma)\times z_i^\prime
\end{equation}
where $\gamma$ is a constant.

\section{Semantic Alignment}
Recent works \cite{li2024imagefolder,wu2024vila} indicate the semantic-rich latent space has substantial benefits to the downstream generation qualities. 

To impose semantics into the tokenized image representation, we propose a semantic alignment term to the quantized token $z^\prime$. A frozen pre-trained \textbf{DINOv2} model \cite{oquab2023dinov2} or \textbf{CLIP} \cite{radford2021learning} can be utilized to extract the semantic-rich visual feature $f$ of the input image $I$. Two types of alignments are supported:
\begin{itemize}
    \item \textbf{Token-level}: The quantized token $z^\prime$ and visual feature $f$ are pooled to $1\times 1$.
    \item \textbf{Patch-level}: The quantized token $z^\prime$ and visual feature $f$ are resized to the same shape.
\end{itemize} 
A CLIP-style contrastive loss \cite{radford2021learning} is adopted to perform visual alignment: maximizing the similarity between the quantized tokens $z^\prime$ and their corresponding DINO representation $f$, while minimizing the similarity between $z^\prime$ and other representations $f$ from different images within one batch. To facilitate semantic learning, we initialize the image encoder $\mathcal{E}$ with the same weight as the frozen one if the XQ-GAN-SC pipeline is leveraged. 

\begin{table*}[t]
\centering
%\captionsetup{justification=raggedleft,singlelinecheck=false}
\scalebox{1}{
\begin{tabular}{
l|p{0.75cm}<{\centering}p{0.75cm}
<{\centering}|p{0.65cm}<{\centering}p{0.65cm}<{\centering}p{0.65cm}<{\centering}p{0.65cm}<{\centering}|p{2.5cm}<{\centering}|p{0.95cm}<{\centering}|p{0.75cm}<{\centering}|p{0.75cm}<{\centering}} 
\hline
Method & RQ & PQ & MS & VQ & LFQ & BSQ & Latent Res. & $|\mathcal{C}|$ & rFID$\downarrow$ & Util.$\uparrow$\\
\hline
\multicolumn{11}{c}{Performance without residual quantization} \\
\hline
VQGAN \cite{esser2021taming} & 1 & 1 & & \checkmark & & & $16\times16$ & 16384 & 4.98 & 63\% \\
VIT-VQGAN \cite{yu2021vector} & 1 & 1 & & \checkmark & & & $16\times16$ & 4096 & 1.55 & 96\% \\
MAGVIT-v2 \cite{yu2024language} & 1 & 1 & & & \checkmark & & $16\times 16$ & 262144 & $\sim$0.9 & 100\% \\
\hline
\rowcolor{gray!10} XQGAN-VP$^2$ & 1 & 2 & & \checkmark & & & $16\times16$ & 4096 & 0.94 & 100\% \\
\rowcolor{gray!10} XQGAN-VP$^2$ & 1 & 2 & & \checkmark & & & $16\times16$ & 16384 & \textbf{0.64} & 100\% \\
\hline
\multicolumn{11}{c}{Performance with residual quantization} \\
\hline
RQGAN \cite{lee2022autoregressiveimagegenerationusing} & 4 & 1 & & \checkmark & & & $16\times16\rightarrow 16\times16$ & 16384 & 1.83 & - \\
VAR \cite{tian2024visualautoregressivemodelingscalable} & 10 & 1 & \checkmark & \checkmark & & & $1\times 1\rightarrow 16\times 16$ & 4096 & 0.90 & 100$\%$ \\
\hline
\rowcolor{gray!10} XQGAN-MSVR$^{10}$P$^2$ & 10 & 2 & \checkmark & \checkmark & & & $1\times1\rightarrow 11\times11$ & 4096 & 0.80 & 100$\%$\\
\rowcolor{gray!10} XQGAN-MSVR$^{10}$P$^2$ & 10 & 2 & \checkmark & \checkmark & & & $1\times1\rightarrow 11\times11$ & 8192 & 0.70 & 100$\%$\\
\rowcolor{gray!10} XQGAN-MSVR$^{10}$P$^2$  & 10 & 2 & \checkmark & \checkmark & & & $1\times1\rightarrow 11\times11$ & 16384 & \textbf{0.67} & 100$\%$\\
\hline
\rowcolor{gray!10} XQGAN-MSBR$^{10}$P$^2$ & 10 & 2 & \checkmark & & & \checkmark & $1\times1\rightarrow 11\times11$ & 4096 & 0.86 & 100$\%$\\
\rowcolor{gray!10} XQGAN-MSBR$^{10}$P$^2$  & 10 & 2 & \checkmark & & & \checkmark & $1\times1\rightarrow 11\times11$ & 16384 & 0.78 & 100$\%$\\
\hline
\end{tabular}
}
\vspace{-0.2cm}
\caption{Reconstruction performance comparison on ImageNet 256x256. $|\mathcal{C}|$ denotes codebook size. $\uparrow$ and $\downarrow$ denote the larger the better and the smaller the better respectively. The XQGAN variants are named by MS(Multi-Scale)-\{V(VQ), L(LFQ), B(BSQ)\}-R$^N$(RQ)-P$^P$(PQ) where $N$ and $P$ denotes residual depth and sub-quantizer number respectively.}
\label{tab:tokenizer}
\vspace{-0.3cm}
\end{table*}

\section{Adversarial Discriminator}
We provide the interface for different discriminators, \ie, PatchGAN, StyleGAN and DINO, different discriminator losses, \ie, hinge, vanilla, and non-saturating, and different generator losses, \ie, hinge and non-saturating.

\paragraph{PatchGAN.}
PatchGAN \cite{isola2017image} is a discriminator architecture commonly used in image tokenizers and generative adversarial networks (GANs). Instead of evaluating the entire image, it operates on smaller patches of an image. This patch-based evaluation encourages the generator to focus on producing fine-grained, localized details that appear realistic at the patch level. By modeling texture, structure, and small-scale details independently of global coherence, PatchGAN effectively guides the generator to improve the quality of individual image regions.

\paragraph{StyleGAN.}
StyleGAN \cite{karras2019style} introduces a more sophisticated discriminator designed to assess image realism while preserving global semantic consistency. Its architecture emphasizes the hierarchical nature of image generation, aligning with StyleGAN's generator that disentangles high-level attributes (\eg, pose and identity) from low-level details (\eg, texture and color). The discriminator aims to ensure that both fine details and overall structure are coherent.

\paragraph{DINOv2.}
DINOv2 \cite{oquab2023dinov2}, as a powerful self-supervised method for learning visual representation, also shows its effectiveness in generating images recently. By leveraging its ability to extract rich semantic features, DINOv2 can be utilized as a discriminator to evaluate the semantic consistency between real and fake images. Compared with PatchGAN \cite{isola2017image} and StyleGAN \cite{karras2019style}, its adversarial is built upon more semantic-rich space and usually achieves a more promising result in reconstruction.

\subsection{Tuning Method}
To facilitate fintuning on pre-trained checkpoints, we support the full-parameter training, LORA \cite{hu2021lora} and frozen for both encoder and decoder.

\subsection{Loss Function} 
The XQGAN is trained with composite losses including reconstruction loss $\mathcal{L}_{recon}$, vector quantization loss $\mathcal{L}_{VQ}$, adverserial loss $\mathcal{L}_{ad}$, Perceptual loss $\mathcal{L}_{P}$, CLIP loss $\mathcal{L}_{clip}$ and auxiliary losses $\mathcal{L}_{aux}$, \eg, entropy loss:
\begin{equation}
\begin{aligned}
\mathcal{L}=\lambda_{recon}\mathcal{L}_{recon} &+ 
    \lambda_{VQ}\mathcal{L}_{VQ} + \lambda_{ad}\mathcal{L}_{ad} + \lambda_{P}\mathcal{L}_{P} + \\
    &\lambda_{clip}\mathcal{L}_{clip}+\lambda_{aux}\mathcal{L}_{aux}.    
\end{aligned}
\end{equation}
Specifically, the reconstruction loss measures the $L_2$ distance between the reconstructed image and the ground truth; vector quantization loss encourages the encoded features and its aligned codebook vectors; adversarial loss, applied from a DINO \cite{oquab2023dinov2} discriminator trained simultaneously, ensures that the generated images are indistinguishable from real ones; perceptual loss compares high-level feature representations from a pre-trained LPIPS \cite{zhang2018unreasonableeffectivenessdeepfeatures} to capture structural differences; and CLIP loss performs semantic regularization between semantic tokens and the pre-trained DINOv2 \cite{oquab2023dinov2} features. A LeCam regularization \cite{tseng2021regularizinggenerativeadversarialnetworks} is applied to the adversarial loss to stabilize the training.

\begin{table*}[]
    \centering
    \begin{tabular}{c|cc|cc|c|cc|c}
    \hline
    ID & Tokenizer & Codebook & Quant.-1 & Quant.-2 & rFID$\downarrow$ & Generator Size & Token Length & gFID$\downarrow$ \\
    \hline
    0 & VAR \cite{tian2024visualautoregressivemodelingscalable} & 4096 & None & - & 0.90 & 310M & 680 & 3.30 \\
    \hline
    1 & XQGAN-MSR$^{10}$P$^2$ & 4096 & DINO & None & 0.80 & 362M & 286 & 2.60 \\
    2 & XQGAN-MSR$^{10}$P$^2$ & 4096 & DINO & SAM & 0.71 & 362M & 286 & 3.46 \\
    \hline
    3 & XQGAN-MSBR$^{10}$P$^2$ & 4096 & None & None & 1.03 & 362M & 286 & 5.83 \\
    4 & XQGAN-MSBR$^{10}$P$^2$ & 4096 & DINO & None & 0.97 & 362M & 286 & 3.54 \\
    5 & XQGAN-MSBR$^{10}$P$^2$ & 4096 & DINO & CLIP & 0.84 & 362M & 286 & 4.39 \\
    6 & XQGAN-MSBR$^{10}$P$^2$ & 4096 & DINO & SAM & 0.86 & 362M & 286 & 3.30 \\
    \hline
    7 & XQGAN-MSBR$^{10}$P$^2$ & 16384 & DINO & SAM & 0.78 & 310M & 286 & 2.96 \\
    \hline
    \end{tabular}
    \vspace{-0.2cm}
    \caption{An overall comparison of XQ-GAN on ImageNet 256x256 with different configurations. Quant.-1 and Quant.-2 denote the alignment models used in the respective quantization branches. All the generators are trained with a 10-step VAR scheme.}
    \vspace{-0.2cm}
    \label{tab:perform}
\end{table*}

\begin{figure*}[t]
    \centering
    \includegraphics[width=0.95\linewidth]{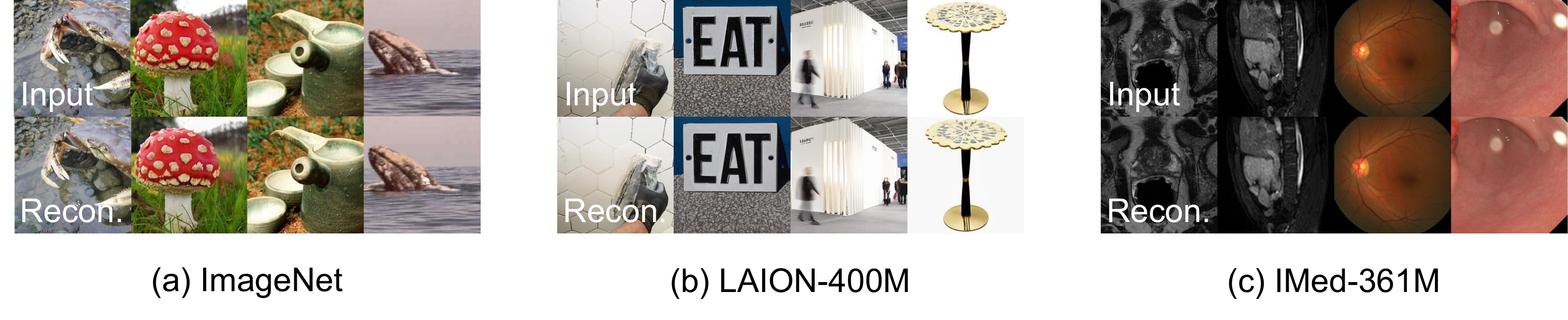}
    \vspace{-0.4cm}
    \caption{Visualization of $256\times 256$ image reconstruction task on (a) Imagenet, (b) LAION-400M, and (c) IMed-361M.}
    \vspace{-0.3cm}
    \label{fig:vis}
\end{figure*}

\section{Experiments}
\subsection{Experimental Settings}
We test our XQ-GAN tokenizer on the ImageNet 256x256 reconstruction and generation tasks. Additional models are trained on a subset of LAION-400M \cite{schuhmann2021laion} (text-rich natural image dataset) and a balanced subset of IMed-361M \cite{cheng2024interactive} (14 categories, multimodal medical image dataset) to provide weights for the community. 

The experiments are conducted on 32 NVIDIA A100 GPUs (80G) or 128 Intel Habana Gaudi HPUs (32G). For the ImageNet dataset, the training of the tokenizer takes about 1.5-2 days depending on the specific setting. The training of the generator takes about 2 days for about 300M VAR \cite{tian2024visualautoregressivemodelingscalable} model (without flash-attention).

\paragraph{Metrics.}
We measure the codebook utilization rate using VAR's implementation \cite{tian2024visualautoregressivemodelingscalable}. For reconstruction and generation, Fréchet Inception Distance (FID) \cite{heusel2017gans} is utilized to evaluate the image quality. 

\paragraph{Implementation details.}
For the XQGAN tokenizer, if there is no other specification, we follow the VQGAN training recipe of LlamaGen \cite{sun2024autoregressive} and use the XQ-GAN-SC pipeline. We set the batch size to 1024 for all experiments. We initialize the image encoder with the weight of the DINOv2-base. We use a cosine learning rate scheduler with a warmup for 1 epoch and a start learning rate of 3e-5. We set the quantizer drop ratio to 0.1. We set $\lambda_{clip}=0.1$, $\lambda_{recon}=\lambda_{VQ}=\lambda_P=1$ and $\lambda_{ad}=0.5$. The residual quantization of each branch shares the same codebook across scales. If quantizer dropout is applied, all branches will share the same dropout schedule. The dropout starts with a $N_{start}=3$ for stability issues. For the generator, we adopt VAR's \cite{tian2024visualautoregressivemodelingscalable} GPT-2-based \cite{radford2019language} architecture.

\begin{figure*}[t]
    \centering
    \includegraphics[width=0.95\linewidth]{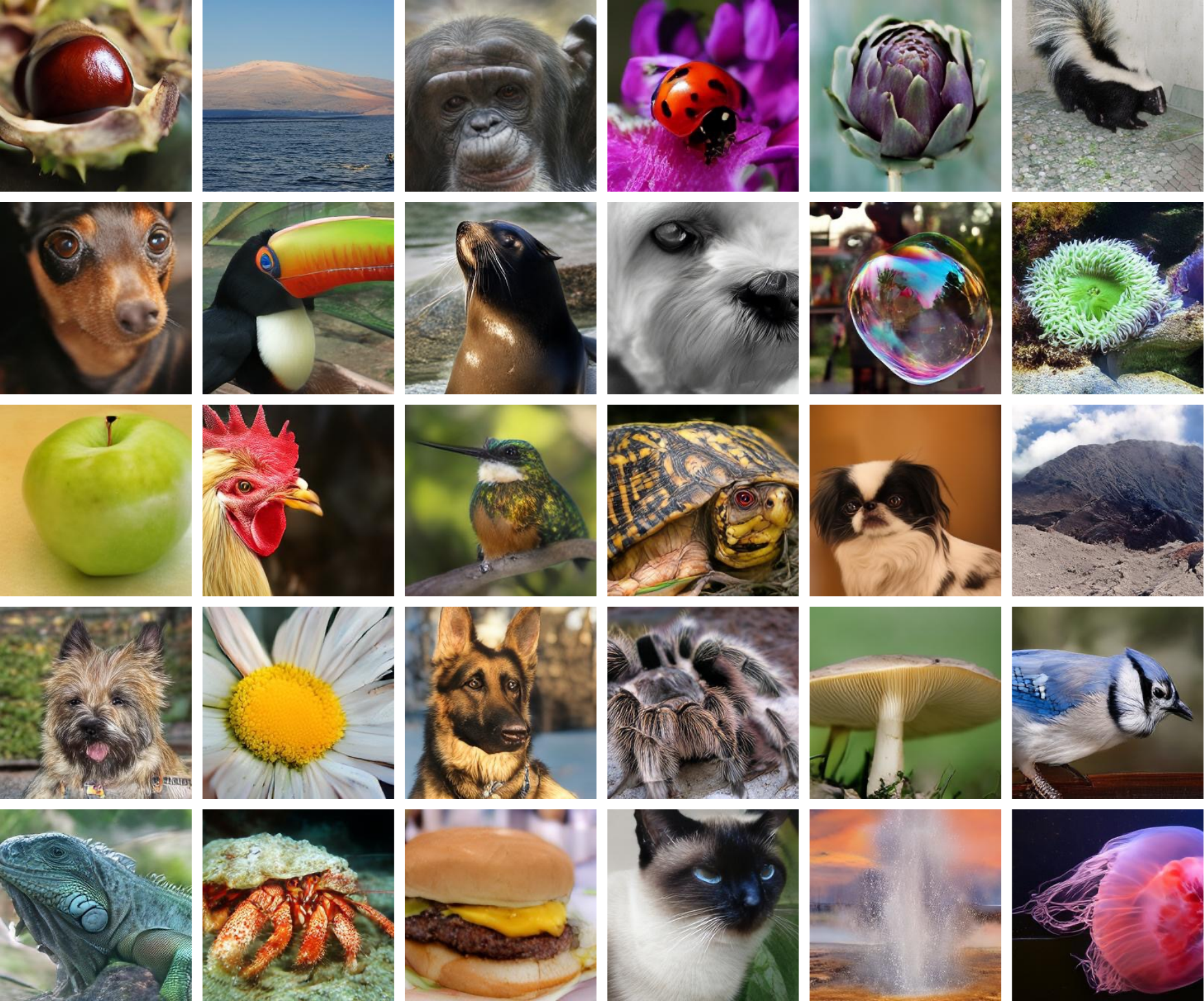}
    \vspace{-0.2cm}
    \caption{Visualization of $256\times 256$ image generation task within ImageNet classes using XQ-GAN-MSVR$^{10}$P$^2$ tokenizer and VAR \cite{tian2024visualautoregressivemodelingscalable} generator.}
    \vspace{-0.3cm}
    \label{fig:vis}
\end{figure*}

\subsection{Main Results on ImageNet}

We summarize our tokenizer's performance on ImageNet reconstruction and generation benchmark of resolution 256x256 in \cref{tab:tokenizer} and \cref{tab:perform} respectively. 

\paragraph{Reconstruction Performance on ImageNet.} To efficiently evaluate the performance of our proposed XQGAN, we compare our method in multiple settings with prior tokenizers. As shown in \cref{tab:tokenizer}, in tokenizers without residual quantization, our XQGAN-VP$^2$ achieves comparable results to MAGViT-v2 \cite{yu2024language} (0.94 rFID vs. 0.90 rFID) despite using a significantly smaller codebook size (4096 vs. 262144). Notably, when the codebook size of XQGAN-VP$^2$ is increased from 4096 to 16384, its performance improves significantly, achieving an rFID of 0.64. On the other hand, for tokenizers with residual quantization, our proposed tokenizer also demonstrates clear advantages over prior approaches. Specifically, XQGAN-MSVR$^{10}$P$^2$ surpasses VAR's tokenizer with a lower rFID (0.86 vs. 0.90) while using significantly fewer tokens (286 vs. 680). This reduction in token count effectively decreases the computational burden during generative model training, improving overall efficiency. To comprehensively evaluate the performance of our tokenizer, we introduce XQGAN-MSBR$^{10}$P$^2$, a novel variant of BSQ. With a codebook size of 16,384, this method achieves an rFID of 0.78, further highlighting the effectiveness of our approach.

\paragraph{Generation Performance on ImageNet.}

Benefiting from our rich-semantic tokenizer, our XQGAN further shows its superior capacity in generation. As shown in \cref{tab:perform}, we evaluate different combinations of alignment models in different quantization branches. (1) In Vector Quantization-based XQGAN, we observe that distilling DINO's \cite{oquab2023dinov2} and SAM's \cite{kirillov2023segany} features into separate branches achieves strong performance (0.71 rFID). However, the best generative performance is obtained when the tokenizer distills only DINO's features within a single quantization branch (2.60 gFID with 0.80 rFID). (2) In Lookup-Free Quantization-based XQGAN, we observe that distilling DINO and SAM features into separate quantization branches achieves the best performance, with an rFID of 0.86 and a gFID of 3.30. Furthermore, XQGAN-MSBR$^{10}$P$^2$ demonstrates promising scalability when increasing the codebook size from 4096 to 16,384, resulting in a 0.34 gFID improvement.

\subsection{Visualization of reconstructed images.}

We present visualizations of reconstructed images from the ImageNet \cite{deng2009imagenet}, LAION-400M \cite{schuhmann2021laion}, and IMed-361M \cite{cheng2024interactive} datasets to further evaluate the performance of our proposed XQGAN. As illustrated in \cref{fig:vis}, our model demonstrates consistent reconstruction quality across diverse datasets and domains, maintaining both semantic accuracy and visual fidelity.

\paragraph{Visualization of generated images on ImageNet.}
We demonstrate qualitative visualization of the generative model trained with XQGAN-MSVR$^{10}$P$^2$ as shown in \cref{fig:vis}. The classes of generated images are among the ImageNet \cite{deng2009imagenet} dataset with a resolution of $256\times 256$.

\section{Conclusion}
In this paper, we comprehensively ablate the image reconstruction and image generation using the XQ-GAN framework. With the hierarchical quantizer design, XQ-GAN achieves state-of-the-art performance for image reconstruction, with latent tokens for both VAR and AR scheme generation. Notably, XQ-GAN is highly customizable and open-source. To facilitate the community's use, we provide pre-trained checkpoints and support multiple fine-tuning settings.

\clearpage

\bibliographystyle{ieeenat_fullname}
\bibliography{main}

\begin{thebibliography}{86}
\providecommand{\natexlab}[1]{#1}
\providecommand{\url}[1]{\texttt{#1}}
\expandafter\ifx\csname urlstyle\endcsname\relax
  \providecommand{\doi}[1]{doi: #1}\else
  \providecommand{\doi}{doi: \begingroup \urlstyle{rm}\Url}\fi

\bibitem[Alexey(2020)]{alexey2020image}
Dosovitskiy Alexey.
\newblock An image is worth 16x16 words: Transformers for image recognition at scale.
\newblock \emph{arXiv preprint arXiv: 2010.11929}, 2020.

\bibitem[Baevski et~al.(2020)Baevski, Zhou, Mohamed, and Auli]{baevski2020wav2vec}
Alexei Baevski, Yuhao Zhou, Abdelrahman Mohamed, and Michael Auli.
\newblock wav2vec 2.0: A framework for self-supervised learning of speech representations.
\newblock \emph{Advances in neural information processing systems}, 33:\penalty0 12449--12460, 2020.

\bibitem[Bao et~al.(2022)Bao, Dong, Piao, and Wei]{bao2022beitbertpretrainingimage}
Hangbo Bao, Li Dong, Songhao Piao, and Furu Wei.
\newblock Beit: Bert pre-training of image transformers, 2022.

\bibitem[Chang et~al.(2022)Chang, Zhang, Jiang, Liu, and Freeman]{chang2022maskgitmaskedgenerativeimage}
Huiwen Chang, Han Zhang, Lu Jiang, Ce Liu, and William~T. Freeman.
\newblock Maskgit: Masked generative image transformer, 2022.

\bibitem[Cheng et~al.(2024)Cheng, Fu, Ye, Wang, Li, Wang, Li, Yao, Chen, Li, et~al.]{cheng2024interactive}
Junlong Cheng, Bin Fu, Jin Ye, Guoan Wang, Tianbin Li, Haoyu Wang, Ruoyu Li, He Yao, Junren Chen, JingWen Li, et~al.
\newblock Interactive medical image segmentation: A benchmark dataset and baseline.
\newblock \emph{arXiv preprint arXiv:2411.12814}, 2024.

\bibitem[Darcet et~al.(2023)Darcet, Oquab, Mairal, and Bojanowski]{darcet2023vitneedreg}
Timothée Darcet, Maxime Oquab, Julien Mairal, and Piotr Bojanowski.
\newblock Vision transformers need registers, 2023.

\bibitem[Deng et~al.(2009)Deng, Dong, Socher, Li, Li, and Fei-Fei]{deng2009imagenet}
Jia Deng, Wei Dong, Richard Socher, Li-Jia Li, Kai Li, and Li Fei-Fei.
\newblock Imagenet: A large-scale hierarchical image database.
\newblock In \emph{2009 IEEE conference on computer vision and pattern recognition}, pages 248--255. Ieee, 2009.

\bibitem[Dhariwal and Nichol(2021)]{dhariwal2021diffusionmodelsbeatgans}
Prafulla Dhariwal and Alex Nichol.
\newblock Diffusion models beat gans on image synthesis, 2021.

\bibitem[Ding et~al.(2021)Ding, Yang, Hong, Zheng, Zhou, Yin, Lin, Zou, Shao, Yang, et~al.]{ding2021cogview}
Ming Ding, Zhuoyi Yang, Wenyi Hong, Wendi Zheng, Chang Zhou, Da Yin, Junyang Lin, Xu Zou, Zhou Shao, Hongxia Yang, et~al.
\newblock Cogview: Mastering text-to-image generation via transformers.
\newblock \emph{Advances in neural information processing systems}, 34:\penalty0 19822--19835, 2021.

\bibitem[Dong et~al.(2023)Dong, Bao, Zhang, Chen, Zhang, Yuan, Chen, Wen, Yu, and Guo]{dong2023peco}
Xiaoyi Dong, Jianmin Bao, Ting Zhang, Dongdong Chen, Weiming Zhang, Lu Yuan, Dong Chen, Fang Wen, Nenghai Yu, and Baining Guo.
\newblock Peco: Perceptual codebook for bert pre-training of vision transformers.
\newblock In \emph{Proceedings of the AAAI Conference on Artificial Intelligence}, pages 552--560, 2023.

\bibitem[Dosovitskiy et~al.(2021)Dosovitskiy, Beyer, Kolesnikov, Weissenborn, Zhai, Unterthiner, Dehghani, Minderer, Heigold, Gelly, Uszkoreit, and Houlsby]{dosovitskiy2021imageworth16x16words}
Alexey Dosovitskiy, Lucas Beyer, Alexander Kolesnikov, Dirk Weissenborn, Xiaohua Zhai, Thomas Unterthiner, Mostafa Dehghani, Matthias Minderer, Georg Heigold, Sylvain Gelly, Jakob Uszkoreit, and Neil Houlsby.
\newblock An image is worth 16x16 words: Transformers for image recognition at scale, 2021.

\bibitem[Esser et~al.(2021)Esser, Rombach, and Ommer]{esser2021taming}
Patrick Esser, Robin Rombach, and Bjorn Ommer.
\newblock Taming transformers for high-resolution image synthesis.
\newblock In \emph{Proceedings of the IEEE/CVF conference on computer vision and pattern recognition}, pages 12873--12883, 2021.

\bibitem[Gafni et~al.(2022)Gafni, Polyak, Ashual, Sheynin, Parikh, and Taigman]{gafni2022make}
Oran Gafni, Adam Polyak, Oron Ashual, Shelly Sheynin, Devi Parikh, and Yaniv Taigman.
\newblock Make-a-scene: Scene-based text-to-image generation with human priors.
\newblock In \emph{European Conference on Computer Vision}, pages 89--106. Springer, 2022.

\bibitem[Gu et~al.(2023)Gu, Wang, Ge, Shan, Qie, and Shou]{gu2023rethinkingobjectivesvectorquantizedtokenizers}
Yuchao Gu, Xintao Wang, Yixiao Ge, Ying Shan, Xiaohu Qie, and Mike~Zheng Shou.
\newblock Rethinking the objectives of vector-quantized tokenizers for image synthesis, 2023.

\bibitem[Hao et~al.(2024)Hao, Ze, Xiang, Ximeng, Fangyi, Jiang, Jindong, Bhiksha, Zicheng, and Emad]{Chen2024softvqvae}
Chen Hao, Wang Ze, Li Xiang, Sun Ximeng, Chen Fangyi, Liu Jiang, Wang Jindong, Raj Bhiksha, Liu Zicheng, and Barsoum Emad.
\newblock Softvq-vae: Efficient 1-dimensional continuous tokenizer.
\newblock \emph{arXiv preprint arXiv}, 2024.

\bibitem[He et~al.(2024)He, Fu, Liu, Wang, Xiao, Shu, Wang, Zhang, Yu, Li, et~al.]{he2024mars}
Wanggui He, Siming Fu, Mushui Liu, Xierui Wang, Wenyi Xiao, Fangxun Shu, Yi Wang, Lei Zhang, Zhelun Yu, Haoyuan Li, et~al.
\newblock Mars: Mixture of auto-regressive models for fine-grained text-to-image synthesis.
\newblock \emph{arXiv preprint arXiv:2407.07614}, 2024.

\bibitem[Heusel et~al.(2017)Heusel, Ramsauer, Unterthiner, Nessler, and Hochreiter]{heusel2017gans}
Martin Heusel, Hubert Ramsauer, Thomas Unterthiner, Bernhard Nessler, and Sepp Hochreiter.
\newblock Gans trained by a two time-scale update rule converge to a local nash equilibrium.
\newblock \emph{Advances in Neural Information Processing Systems}, 30, 2017.

\bibitem[Hinton and Salakhutdinov(2006)]{hinton2006reducing}
Geoffrey~E Hinton and Ruslan~R Salakhutdinov.
\newblock Reducing the dimensionality of data with neural networks.
\newblock \emph{science}, 313\penalty0 (5786):\penalty0 504--507, 2006.

\bibitem[Hu et~al.(2021)Hu, Shen, Wallis, Allen-Zhu, Li, Wang, and Chen]{hu2021lora}
Edward~J. Hu, Yelong Shen, Phillip Wallis, Zeyuan Allen-Zhu, Yuanzhi Li, Shean Wang, and Weizhu Chen.
\newblock Lora: Low-rank adaptation of large language models, 2021.

\bibitem[Huang et~al.(2023)Huang, Mao, Chen, and Zhang]{huang2023towards}
Mengqi Huang, Zhendong Mao, Zhuowei Chen, and Yongdong Zhang.
\newblock Towards accurate image coding: Improved autoregressive image generation with dynamic vector quantization.
\newblock In \emph{Proceedings of the IEEE/CVF Conference on Computer Vision and Pattern Recognition}, pages 22596--22605, 2023.

\bibitem[Isola et~al.(2017)Isola, Zhu, Zhou, and Efros]{isola2017image}
Phillip Isola, Jun-Yan Zhu, Tinghui Zhou, and Alexei~A. Efros.
\newblock Image-to-image translation with conditional adversarial networks.
\newblock In \emph{Proceedings of the IEEE Conference on Computer Vision and Pattern Recognition (CVPR)}, pages 1125--1134, 2017.

\bibitem[Karras et~al.(2019)Karras, Laine, and Aila]{karras2019style}
Tero Karras, Samuli Laine, and Timo Aila.
\newblock A style-based generator architecture for generative adversarial networks.
\newblock In \emph{Proceedings of the IEEE/CVF Conference on Computer Vision and Pattern Recognition (CVPR)}, pages 4401--4410, 2019.

\bibitem[Kirillov et~al.(2023)Kirillov, Mintun, Ravi, Mao, Rolland, Gustafson, Xiao, Whitehead, Berg, Lo, Doll{\'a}r, and Girshick]{kirillov2023segany}
Alexander Kirillov, Eric Mintun, Nikhila Ravi, Hanzi Mao, Chloe Rolland, Laura Gustafson, Tete Xiao, Spencer Whitehead, Alexander~C. Berg, Wan-Yen Lo, Piotr Doll{\'a}r, and Ross Girshick.
\newblock Segment anything.
\newblock \emph{arXiv:2304.02643}, 2023.

\bibitem[Kumar et~al.(2024)Kumar, Seetharaman, Luebs, Kumar, and Kumar]{kumar2024high}
Rithesh Kumar, Prem Seetharaman, Alejandro Luebs, Ishaan Kumar, and Kundan Kumar.
\newblock High-fidelity audio compression with improved rvqgan.
\newblock \emph{Advances in Neural Information Processing Systems}, 36, 2024.

\bibitem[Lee et~al.(2022{\natexlab{a}})Lee, Kim, Kim, Cho, and Han]{lee2022autoregressive}
Doyup Lee, Chiheon Kim, Saehoon Kim, Minsu Cho, and Wook-Shin Han.
\newblock Autoregressive image generation using residual quantization.
\newblock In \emph{Proceedings of the IEEE/CVF Conference on Computer Vision and Pattern Recognition}, pages 11523--11532, 2022{\natexlab{a}}.

\bibitem[Lee et~al.(2022{\natexlab{b}})Lee, Kim, Kim, Cho, and Han]{lee2022autoregressiveimagegenerationusing}
Doyup Lee, Chiheon Kim, Saehoon Kim, Minsu Cho, and Wook-Shin Han.
\newblock Autoregressive image generation using residual quantization, 2022{\natexlab{b}}.

\bibitem[Li et~al.(2024{\natexlab{a}})Li, Yang, Wang, Qiu, Chou, Li, and Li]{li2024scalable}
Haopeng Li, Jinyue Yang, Kexin Wang, Xuerui Qiu, Yuhong Chou, Xin Li, and Guoqi Li.
\newblock Scalable autoregressive image generation with mamba.
\newblock \emph{arXiv preprint arXiv:2408.12245}, 2024{\natexlab{a}}.

\bibitem[Li et~al.(2023)Li, Chang, Mishra, Zhang, Katabi, and Krishnan]{li2023magemaskedgenerativeencoder}
Tianhong Li, Huiwen Chang, Shlok~Kumar Mishra, Han Zhang, Dina Katabi, and Dilip Krishnan.
\newblock Mage: Masked generative encoder to unify representation learning and image synthesis, 2023.

\bibitem[Li et~al.(2024{\natexlab{b}})Li, Tian, Li, Deng, and He]{li2024autoregressiveimagegenerationvector}
Tianhong Li, Yonglong Tian, He Li, Mingyang Deng, and Kaiming He.
\newblock Autoregressive image generation without vector quantization, 2024{\natexlab{b}}.

\bibitem[Li et~al.(2024{\natexlab{c}})Li, Qiu, Chen, Kuen, Gu, Raj, and Lin]{li2024imagefolder}
Xiang Li, Kai Qiu, Hao Chen, Jason Kuen, Jiuxiang Gu, Bhiksha Raj, and Zhe Lin.
\newblock Imagefolder: Autoregressive image generation with folded tokens.
\newblock \emph{arXiv preprint arXiv:2410.01756}, 2024{\natexlab{c}}.

\bibitem[Li et~al.(2024{\natexlab{d}})Li, Qiu, Chen, Kuen, Lin, Singh, and Raj]{li2024controlvar}
Xiang Li, Kai Qiu, Hao Chen, Jason Kuen, Zhe Lin, Rita Singh, and Bhiksha Raj.
\newblock Controlvar: Exploring controllable visual autoregressive modeling.
\newblock \emph{arXiv preprint arXiv:2406.09750}, 2024{\natexlab{d}}.

\bibitem[Liu et~al.(2024)Liu, Zhao, Zhuo, Lin, Qiao, Li, and Gao]{liu2024lumina}
Dongyang Liu, Shitian Zhao, Le Zhuo, Weifeng Lin, Yu Qiao, Hongsheng Li, and Peng Gao.
\newblock Lumina-mgpt: Illuminate flexible photorealistic text-to-image generation with multimodal generative pretraining.
\newblock \emph{arXiv preprint arXiv:2408.02657}, 2024.

\bibitem[Luo et~al.(2024)Luo, Shi, Ge, Yang, Wang, and Shan]{luo2024open}
Zhuoyan Luo, Fengyuan Shi, Yixiao Ge, Yujiu Yang, Limin Wang, and Ying Shan.
\newblock Open-magvit2: An open-source project toward democratizing auto-regressive visual generation.
\newblock \emph{arXiv preprint arXiv:2409.04410}, 2024.

\bibitem[Mentzer et~al.(2023)Mentzer, Minnen, Agustsson, and Tschannen]{mentzer2023finite}
Fabian Mentzer, David Minnen, Eirikur Agustsson, and Michael Tschannen.
\newblock Finite scalar quantization: Vq-vae made simple, 2023.

\bibitem[Mizrahi et~al.(2024)Mizrahi, Bachmann, Kar, Yeo, Gao, Dehghan, and Zamir]{mizrahi20244m}
David Mizrahi, Roman Bachmann, Oguzhan Kar, Teresa Yeo, Mingfei Gao, Afshin Dehghan, and Amir Zamir.
\newblock 4m: Massively multimodal masked modeling.
\newblock \emph{Advances in Neural Information Processing Systems}, 36, 2024.

\bibitem[Nichol and Dhariwal(2021)]{nichol2021improveddenoisingdiffusionprobabilistic}
Alex Nichol and Prafulla Dhariwal.
\newblock Improved denoising diffusion probabilistic models, 2021.

\bibitem[Nichol et~al.(2021)Nichol, Dhariwal, Ramesh, Shyam, Mishkin, McGrew, Sutskever, and Chen]{nichol2021glide}
Alex Nichol, Prafulla Dhariwal, Aditya Ramesh, Pranav Shyam, Pamela Mishkin, Bob McGrew, Ilya Sutskever, and Mark Chen.
\newblock Glide: Towards photorealistic image generation and editing with text-guided diffusion models.
\newblock \emph{arXiv preprint arXiv:2112.10741}, 2021.

\bibitem[Ning et~al.()Ning, Li, Zhang, Geng, Dai, He, and Hu]{ning2301all}
J Ning, C Li, Z Zhang, Z Geng, Q Dai, K He, and H Hu.
\newblock All in tokens: Unifying output space of visual tasks via soft token. arxiv 2023.
\newblock \emph{arXiv preprint arXiv:2301.02229}.

\bibitem[Oquab et~al.(2023)Oquab, Darcet, Moutakanni, Vo, Szafraniec, Khalidov, Fernandez, Haziza, Massa, El-Nouby, Howes, Huang, Xu, Sharma, Li, Galuba, Rabbat, Assran, Ballas, Synnaeve, Misra, Jegou, Mairal, Labatut, Joulin, and Bojanowski]{oquab2023dinov2}
Maxime Oquab, Timothée Darcet, Theo Moutakanni, Huy~V. Vo, Marc Szafraniec, Vasil Khalidov, Pierre Fernandez, Daniel Haziza, Francisco Massa, Alaaeldin El-Nouby, Russell Howes, Po-Yao Huang, Hu Xu, Vasu Sharma, Shang-Wen Li, Wojciech Galuba, Mike Rabbat, Mido Assran, Nicolas Ballas, Gabriel Synnaeve, Ishan Misra, Herve Jegou, Julien Mairal, Patrick Labatut, Armand Joulin, and Piotr Bojanowski.
\newblock Dinov2: Learning robust visual features without supervision, 2023.

\bibitem[Peebles and Xie(2023)]{peebles2023scalablediffusionmodelstransformers}
William Peebles and Saining Xie.
\newblock Scalable diffusion models with transformers, 2023.

\bibitem[Peng et~al.(2022)Peng, Dong, Bao, Ye, and Wei]{peng2022beitv2maskedimage}
Zhiliang Peng, Li Dong, Hangbo Bao, Qixiang Ye, and Furu Wei.
\newblock Beit v2: Masked image modeling with vector-quantized visual tokenizers, 2022.

\bibitem[Radford et~al.(2019)Radford, Wu, Child, Luan, Amodei, Sutskever, et~al.]{radford2019language}
Alec Radford, Jeffrey Wu, Rewon Child, David Luan, Dario Amodei, Ilya Sutskever, et~al.
\newblock Language models are unsupervised multitask learners.
\newblock \emph{OpenAI blog}, 1\penalty0 (8):\penalty0 9, 2019.

\bibitem[Radford et~al.(2021)Radford, Kim, Hallacy, Ramesh, Goh, Agarwal, Sastry, Askell, Mishkin, Clark, et~al.]{radford2021learning}
Alec Radford, Jong~Wook Kim, Chris Hallacy, Aditya Ramesh, Gabriel Goh, Sandhini Agarwal, Girish Sastry, Amanda Askell, Pamela Mishkin, Jack Clark, et~al.
\newblock Learning transferable visual models from natural language supervision.
\newblock In \emph{International conference on machine learning}, pages 8748--8763. PMLR, 2021.

\bibitem[Ramesh et~al.(2021)Ramesh, Pavlov, Goh, Gray, Voss, Radford, Chen, and Sutskever]{ramesh2021zero}
Aditya Ramesh, Mikhail Pavlov, Gabriel Goh, Scott Gray, Chelsea Voss, Alec Radford, Mark Chen, and Ilya Sutskever.
\newblock Zero-shot text-to-image generation.
\newblock In \emph{International conference on machine learning}, pages 8821--8831. Pmlr, 2021.

\bibitem[Razavi et~al.(2019{\natexlab{a}})Razavi, Van~den Oord, and Vinyals]{razavi2019generating}
Ali Razavi, Aaron Van~den Oord, and Oriol Vinyals.
\newblock Generating diverse high-fidelity images with vq-vae-2.
\newblock \emph{Advances in neural information processing systems}, 32, 2019{\natexlab{a}}.

\bibitem[Razavi et~al.(2019{\natexlab{b}})Razavi, van~den Oord, and Vinyals]{razavi2019generatingdiversehighfidelityimages}
Ali Razavi, Aaron van~den Oord, and Oriol Vinyals.
\newblock Generating diverse high-fidelity images with vq-vae-2, 2019{\natexlab{b}}.

\bibitem[Rombach et~al.(2022{\natexlab{a}})Rombach, Blattmann, Lorenz, Esser, and Ommer]{rombach2022high}
Robin Rombach, Andreas Blattmann, Dominik Lorenz, Patrick Esser, and Bj{\"o}rn Ommer.
\newblock High-resolution image synthesis with latent diffusion models.
\newblock In \emph{Proceedings of the IEEE/CVF conference on computer vision and pattern recognition}, pages 10684--10695, 2022{\natexlab{a}}.

\bibitem[Rombach et~al.(2022{\natexlab{b}})Rombach, Blattmann, Lorenz, Esser, and Ommer]{rombach2022highresolutionimagesynthesislatent}
Robin Rombach, Andreas Blattmann, Dominik Lorenz, Patrick Esser, and Björn Ommer.
\newblock High-resolution image synthesis with latent diffusion models, 2022{\natexlab{b}}.

\bibitem[Ronneberger et~al.(2015)Ronneberger, Fischer, and Brox]{ronneberger2015u}
Olaf Ronneberger, Philipp Fischer, and Thomas Brox.
\newblock U-net: Convolutional networks for biomedical image segmentation.
\newblock In \emph{Medical image computing and computer-assisted intervention--MICCAI 2015: 18th international conference, Munich, Germany, October 5-9, 2015, proceedings, part III 18}, pages 234--241. Springer, 2015.

\bibitem[Saharia et~al.(2022)Saharia, Chan, Saxena, Li, Whang, Denton, Ghasemipour, Gontijo~Lopes, Karagol~Ayan, Salimans, et~al.]{saharia2022photorealistic}
Chitwan Saharia, William Chan, Saurabh Saxena, Lala Li, Jay Whang, Emily~L Denton, Kamyar Ghasemipour, Raphael Gontijo~Lopes, Burcu Karagol~Ayan, Tim Salimans, et~al.
\newblock Photorealistic text-to-image diffusion models with deep language understanding.
\newblock \emph{Advances in neural information processing systems}, 35:\penalty0 36479--36494, 2022.

\bibitem[Schuhmann et~al.(2021)Schuhmann, Vencu, Beaumont, Kaczmarczyk, Mullis, Katta, Coombes, Jitsev, and Komatsuzaki]{schuhmann2021laion}
Christoph Schuhmann, Richard Vencu, Romain Beaumont, Robert Kaczmarczyk, Clayton Mullis, Aarush Katta, Theo Coombes, Jenia Jitsev, and Aran Komatsuzaki.
\newblock Laion-400m: Open dataset of clip-filtered 400 million image-text pairs.
\newblock \emph{arXiv preprint arXiv:2111.02114}, 2021.

\bibitem[Shi et~al.(2022)Shi, Wu, Liang, Liu, and Duan]{shi2022divaephotorealisticimagessynthesis}
Jie Shi, Chenfei Wu, Jian Liang, Xiang Liu, and Nan Duan.
\newblock Divae: Photorealistic images synthesis with denoising diffusion decoder, 2022.

\bibitem[Sohl-Dickstein et~al.(2015)Sohl-Dickstein, Weiss, Maheswaranathan, and Ganguli]{sohldickstein2015deepunsupervisedlearningusing}
Jascha Sohl-Dickstein, Eric~A. Weiss, Niru Maheswaranathan, and Surya Ganguli.
\newblock Deep unsupervised learning using nonequilibrium thermodynamics, 2015.

\bibitem[Song et~al.(2022)Song, Meng, and Ermon]{song2022denoisingdiffusionimplicitmodels}
Jiaming Song, Chenlin Meng, and Stefano Ermon.
\newblock Denoising diffusion implicit models, 2022.

\bibitem[Sun et~al.(2024)Sun, Jiang, Chen, Zhang, Peng, Luo, and Yuan]{sun2024autoregressive}
Peize Sun, Yi Jiang, Shoufa Chen, Shilong Zhang, Bingyue Peng, Ping Luo, and Zehuan Yuan.
\newblock Autoregressive model beats diffusion: Llama for scalable image generation.
\newblock \emph{arXiv preprint arXiv:2406.06525}, 2024.

\bibitem[Takida et~al.(2023)Takida, Ikemiya, Shibuya, Shimada, Choi, Lai, Murata, Uesaka, Uchida, Liao, et~al.]{takida2023hq}
Yuhta Takida, Yukara Ikemiya, Takashi Shibuya, Kazuki Shimada, Woosung Choi, Chieh-Hsin Lai, Naoki Murata, Toshimitsu Uesaka, Kengo Uchida, Wei-Hsiang Liao, et~al.
\newblock Hq-vae: Hierarchical discrete representation learning with variational bayes.
\newblock \emph{arXiv preprint arXiv:2401.00365}, 2023.

\bibitem[Tian et~al.(2024)Tian, Jiang, Yuan, Peng, and Wang]{tian2024visualautoregressivemodelingscalable}
Keyu Tian, Yi Jiang, Zehuan Yuan, Bingyue Peng, and Liwei Wang.
\newblock Visual autoregressive modeling: Scalable image generation via next-scale prediction, 2024.

\bibitem[Touvron et~al.(2023)Touvron, Lavril, Izacard, Martinet, Lachaux, Lacroix, Rozière, Goyal, Hambro, Azhar, Rodriguez, Joulin, Grave, and Lample]{touvron2023llamaopenefficientfoundation}
Hugo Touvron, Thibaut Lavril, Gautier Izacard, Xavier Martinet, Marie-Anne Lachaux, Timothée Lacroix, Baptiste Rozière, Naman Goyal, Eric Hambro, Faisal Azhar, Aurelien Rodriguez, Armand Joulin, Edouard Grave, and Guillaume Lample.
\newblock Llama: Open and efficient foundation language models, 2023.

\bibitem[Tseng et~al.(2021)Tseng, Jiang, Liu, Yang, and Yang]{tseng2021regularizinggenerativeadversarialnetworks}
Hung-Yu Tseng, Lu Jiang, Ce Liu, Ming-Hsuan Yang, and Weilong Yang.
\newblock Regularizing generative adversarial networks under limited data, 2021.

\bibitem[Vahdat et~al.(2021)Vahdat, Kreis, and Kautz]{vahdat2021scorebasedgenerativemodelinglatent}
Arash Vahdat, Karsten Kreis, and Jan Kautz.
\newblock Score-based generative modeling in latent space, 2021.

\bibitem[Van~den Oord et~al.(2016)Van~den Oord, Kalchbrenner, Espeholt, Vinyals, Graves, et~al.]{van2016conditional}
Aaron Van~den Oord, Nal Kalchbrenner, Lasse Espeholt, Oriol Vinyals, Alex Graves, et~al.
\newblock Conditional image generation with pixelcnn decoders.
\newblock \emph{Advances in neural information processing systems}, 29, 2016.

\bibitem[Van Den~Oord et~al.(2017)Van Den~Oord, Vinyals, et~al.]{van2017neural}
Aaron Van Den~Oord, Oriol Vinyals, et~al.
\newblock Neural discrete representation learning.
\newblock \emph{Advances in neural information processing systems}, 30, 2017.

\bibitem[Vaswani et~al.(2023)Vaswani, Shazeer, Parmar, Uszkoreit, Jones, Gomez, Kaiser, and Polosukhin]{vaswani2023attentionneed}
Ashish Vaswani, Noam Shazeer, Niki Parmar, Jakob Uszkoreit, Llion Jones, Aidan~N. Gomez, Lukasz Kaiser, and Illia Polosukhin.
\newblock Attention is all you need, 2023.

\bibitem[Vincent et~al.(2008)Vincent, Larochelle, Bengio, and Manzagol]{vincent2008extracting}
Pascal Vincent, Hugo Larochelle, Yoshua Bengio, and Pierre-Antoine Manzagol.
\newblock Extracting and composing robust features with denoising autoencoders.
\newblock In \emph{Proceedings of the 25th international conference on Machine learning}, pages 1096--1103, 2008.

\bibitem[Wang et~al.(2021)Wang, Zhu, Adam, Yuille, and Chen]{wang2021maxdeeplabendtoendpanopticsegmentation}
Huiyu Wang, Yukun Zhu, Hartwig Adam, Alan Yuille, and Liang-Chieh Chen.
\newblock Max-deeplab: End-to-end panoptic segmentation with mask transformers, 2021.

\bibitem[Weber et~al.(2024)Weber, Yu, Yu, Deng, Shen, Cremers, and Chen]{weber2024maskbit}
Mark Weber, Lijun Yu, Qihang Yu, Xueqing Deng, Xiaohui Shen, Daniel Cremers, and Liang-Chieh Chen.
\newblock Maskbit: Embedding-free image generation via bit tokens.
\newblock \emph{arXiv preprint arXiv:2409.16211}, 2024.

\bibitem[Wu et~al.(2024)Wu, Zhang, Chen, Tang, Li, Fang, Zhu, Xie, Yin, Yi, et~al.]{wu2024vila}
Yecheng Wu, Zhuoyang Zhang, Junyu Chen, Haotian Tang, Dacheng Li, Yunhao Fang, Ligeng Zhu, Enze Xie, Hongxu Yin, Li Yi, et~al.
\newblock Vila-u: a unified foundation model integrating visual understanding and generation.
\newblock \emph{arXiv preprint arXiv:2409.04429}, 2024.

\bibitem[Xie et~al.(2024)Xie, Mao, Bai, Zhang, Wang, Lin, Gu, Chen, Yang, and Shou]{xie2024show}
Jinheng Xie, Weijia Mao, Zechen Bai, David~Junhao Zhang, Weihao Wang, Kevin~Qinghong Lin, Yuchao Gu, Zhijie Chen, Zhenheng Yang, and Mike~Zheng Shou.
\newblock Show-o: One single transformer to unify multimodal understanding and generation.
\newblock \emph{arXiv preprint arXiv:2408.12528}, 2024.

\bibitem[Yang et~al.(2024)Yang, Kang, Huang, Xu, Feng, and Zhao]{yang2024depth}
Lihe Yang, Bingyi Kang, Zilong Huang, Xiaogang Xu, Jiashi Feng, and Hengshuang Zhao.
\newblock Depth anything: Unleashing the power of large-scale unlabeled data.
\newblock In \emph{Proceedings of the IEEE/CVF Conference on Computer Vision and Pattern Recognition}, pages 10371--10381, 2024.

\bibitem[Yu et~al.(2021)Yu, Li, Koh, Zhang, Pang, Qin, Ku, Xu, Baldridge, and Wu]{yu2021vector}
Jiahui Yu, Xin Li, Jing~Yu Koh, Han Zhang, Ruoming Pang, James Qin, Alexander Ku, Yuanzhong Xu, Jason Baldridge, and Yonghui Wu.
\newblock Vector-quantized image modeling with improved vqgan.
\newblock \emph{arXiv preprint arXiv:2110.04627}, 2021.

\bibitem[Yu et~al.(2023{\natexlab{a}})Yu, Cheng, Sohn, Lezama, Zhang, Chang, Hauptmann, Yang, Hao, Essa, et~al.]{yu2023magvit}
Lijun Yu, Yong Cheng, Kihyuk Sohn, Jos{\'e} Lezama, Han Zhang, Huiwen Chang, Alexander~G Hauptmann, Ming-Hsuan Yang, Yuan Hao, Irfan Essa, et~al.
\newblock Magvit: Masked generative video transformer.
\newblock In \emph{Proceedings of the IEEE/CVF Conference on Computer Vision and Pattern Recognition}, pages 10459--10469, 2023{\natexlab{a}}.

\bibitem[Yu et~al.(2023{\natexlab{b}})Yu, Lezama, Gundavarapu, Versari, Sohn, Minnen, Cheng, Gupta, Gu, Hauptmann, Gong, Yang, Essa, Ross, and Jiang]{yu2023language}
Lijun Yu, José Lezama, Nitesh~B. Gundavarapu, Luca Versari, Kihyuk Sohn, David Minnen, Yong Cheng, Agrim Gupta, Xiuye Gu, Alexander~G. Hauptmann, Boqing Gong, Ming-Hsuan Yang, Irfan Essa, David~A. Ross, and Lu Jiang.
\newblock Language model beats diffusion -- tokenizer is key to visual generation, 2023{\natexlab{b}}.

\bibitem[Yu et~al.(2024{\natexlab{a}})Yu, Cheng, Wang, Kumar, Macherey, Huang, Ross, Essa, Bisk, Yang, et~al.]{yu2024spae}
Lijun Yu, Yong Cheng, Zhiruo Wang, Vivek Kumar, Wolfgang Macherey, Yanping Huang, David Ross, Irfan Essa, Yonatan Bisk, Ming-Hsuan Yang, et~al.
\newblock Spae: Semantic pyramid autoencoder for multimodal generation with frozen llms.
\newblock \emph{Advances in Neural Information Processing Systems}, 36, 2024{\natexlab{a}}.

\bibitem[Yu et~al.(2024{\natexlab{b}})Yu, Lezama, Gundavarapu, Versari, Sohn, Minnen, Cheng, Gupta, Gu, Hauptmann, Gong, Yang, Essa, Ross, and Jiang]{yu2024language}
Lijun Yu, Jose Lezama, Nitesh~Bharadwaj Gundavarapu, Luca Versari, Kihyuk Sohn, David Minnen, Yong Cheng, Agrim Gupta, Xiuye Gu, Alexander~G Hauptmann, Boqing Gong, Ming-Hsuan Yang, Irfan Essa, David~A Ross, and Lu Jiang.
\newblock Language model beats diffusion - tokenizer is key to visual generation.
\newblock In \emph{The Twelfth International Conference on Learning Representations}, 2024{\natexlab{b}}.

\bibitem[Yu et~al.(2024{\natexlab{c}})Yu, Weber, Deng, Shen, Cremers, and Chen]{yu2024an}
Qihang Yu, Mark Weber, Xueqing Deng, Xiaohui Shen, Daniel Cremers, and Liang-Chieh Chen.
\newblock An image is worth 32 tokens for reconstruction and generation.
\newblock \emph{arxiv: 2406.07550}, 2024{\natexlab{c}}.

\bibitem[Yu et~al.(2024{\natexlab{d}})Yu, Weber, Deng, Shen, Cremers, and Chen]{yu2024imageworth32tokens}
Qihang Yu, Mark Weber, Xueqing Deng, Xiaohui Shen, Daniel Cremers, and Liang-Chieh Chen.
\newblock An image is worth 32 tokens for reconstruction and generation, 2024{\natexlab{d}}.

\bibitem[Yu et~al.(2024{\natexlab{e}})Yu, Kwak, Jang, Jeong, Huang, Shin, and Xie]{yu2024representation}
Sihyun Yu, Sangkyung Kwak, Huiwon Jang, Jongheon Jeong, Jonathan Huang, Jinwoo Shin, and Saining Xie.
\newblock Representation alignment for generation: Training diffusion transformers is easier than you think.
\newblock \emph{arXiv preprint arXiv:2410.06940}, 2024{\natexlab{e}}.

\bibitem[Zeghidour et~al.(2021)Zeghidour, Luebs, Omran, Skoglund, and Tagliasacchi]{zeghidour2021soundstream}
Neil Zeghidour, Alejandro Luebs, Ahmed Omran, Jan Skoglund, and Marco Tagliasacchi.
\newblock Soundstream: An end-to-end neural audio codec.
\newblock \emph{IEEE/ACM Transactions on Audio, Speech, and Language Processing}, 30:\penalty0 495--507, 2021.

\bibitem[Zhang et~al.(2018)Zhang, Isola, Efros, Shechtman, and Wang]{zhang2018unreasonableeffectivenessdeepfeatures}
Richard Zhang, Phillip Isola, Alexei~A. Efros, Eli Shechtman, and Oliver Wang.
\newblock The unreasonable effectiveness of deep features as a perceptual metric, 2018.

\bibitem[Zhao et~al.(2024)Zhao, Xiong, and Kr{\"a}henb{\"u}hl]{zhao2024image}
Yue Zhao, Yuanjun Xiong, and Philipp Kr{\"a}henb{\"u}hl.
\newblock Image and video tokenization with binary spherical quantization.
\newblock \emph{arXiv preprint arXiv:2406.07548}, 2024.

\bibitem[Zheng et~al.(2022)Zheng, Vuong, Cai, and Phung]{zheng2022movqmodulatingquantizedvectors}
Chuanxia Zheng, Long~Tung Vuong, Jianfei Cai, and Dinh Phung.
\newblock Movq: Modulating quantized vectors for high-fidelity image generation, 2022.

\bibitem[Zhou et~al.(2024)Zhou, Yu, Babu, Tirumala, Yasunaga, Shamis, Kahn, Ma, Zettlemoyer, and Levy]{zhou2024transfusion}
Chunting Zhou, Lili Yu, Arun Babu, Kushal Tirumala, Michihiro Yasunaga, Leonid Shamis, Jacob Kahn, Xuezhe Ma, Luke Zettlemoyer, and Omer Levy.
\newblock Transfusion: Predict the next token and diffuse images with one multi-modal model.
\newblock \emph{arXiv preprint arXiv:2408.11039}, 2024.

\bibitem[Zhu et~al.(2024{\natexlab{a}})Zhu, Wei, Lu, and Chen]{zhu2024scaling}
Lei Zhu, Fangyun Wei, Yanye Lu, and Dong Chen.
\newblock Scaling the codebook size of vqgan to 100,000 with a utilization rate of 99\%.
\newblock \emph{arXiv preprint arXiv:2406.11837}, 2024{\natexlab{a}}.

\bibitem[Zhu et~al.(2010)Zhu, Su, Lu, Li, Wang, and Dai]{zhu2010deformable}
X Zhu, W Su, L Lu, B Li, X Wang, and J Dai.
\newblock Deformable detr: Deformable transformers for end-to-end object detection. arxiv 2020.
\newblock \emph{arXiv preprint arXiv:2010.04159}, 2010.

\bibitem[Zhu et~al.(2024{\natexlab{b}})Zhu, Li, Xin, and Xu]{zhu2024addressing}
Yongxin Zhu, Bocheng Li, Yifei Xin, and Linli Xu.
\newblock Addressing representation collapse in vector quantized models with one linear layer.
\newblock \emph{arXiv preprint arXiv:2411.02038}, 2024{\natexlab{b}}.

\bibitem[Zhu et~al.(2024{\natexlab{c}})Zhu, Li, Zhang, Li, Xu, and Bing]{zhu2024stabilize}
Yongxin Zhu, Bocheng Li, Hang Zhang, Xin Li, Linli Xu, and Lidong Bing.
\newblock Stabilize the latent space for image autoregressive modeling: A unified perspective.
\newblock \emph{arXiv preprint arXiv:2410.12490}, 2024{\natexlab{c}}.

\end{thebibliography}

% \renewcommand{\thetable}{{\Alph{table}}}
% \renewcommand{\thefigure}{{\Alph{figure}}}
% \renewcommand{\thesection}{{\Alph{section}}}
% \setcounter{figure}{0}   
% \setcounter{table}{0}   
% \setcounter{section}{0} 

% \clearpage
% \input{src/7-appendix}

% \clearpage
% \appendix
% \input{src/7-appendix}

% \clearpage
% \input{src/6-checklist}

\end{document}